\documentclass[10pt,twocolumn,letterpaper]{article}

\usepackage{iccv}
\usepackage{times}
\usepackage{epsfig}
\usepackage{graphicx}
\usepackage{amsmath}
\usepackage{amssymb}

\usepackage{algorithm}
\usepackage{bm}
\usepackage{algorithmic}
\usepackage{listings}
\usepackage{makecell}  
\usepackage{diagbox} 
\usepackage{mathtools} 
\usepackage{multirow}
\usepackage{pifont}
\newcommand{\cmark}{\ding{51}}%
\newcommand{\xmark}{\ding{55}}%
\usepackage[table]{xcolor}
\usepackage[pagebackref, breaklinks=true, colorlinks, citecolor=citecolor, linkcolor=linkcolor, bookmarks=false]{hyperref}
\definecolor{citecolor}{HTML}{0071BC}
\definecolor{linkcolor}{HTML}{ED1C24}

\definecolor{detcolor}{gray}{.9}

\definecolor{bestcolor}{gray}{.9}

\newlength\savewidth
  
\newcolumntype{x}[1]{>{\centering\arraybackslash}p{#1pt}}
\newcolumntype{y}[1]{>{\raggedright\arraybackslash}p{#1pt}}
\newcolumntype{z}[1]{>{\raggedleft\arraybackslash}p{#1pt}}
\renewcommand{\paragraph}[1]{\vspace{1.25mm}\noindent\textbf{#1}}
\definecolor{deemph}{gray}{0.6}

\usepackage{etoolbox}
\makeatletter
\AfterEndEnvironment{algorithm}{\let\@algcomment\relax}
\AtEndEnvironment{algorithm}{\kern2pt\hrule\relax\vskip3pt\@algcomment}
\let\@algcomment\relax
\newcommand\algcomment[1]{\def\@algcomment{\footnotesize#1}}
\renewcommand\fs@ruled{\def\@fs@cfont{\bfseries}\let\@fs@capt\floatc@ruled
  \def\@fs@pre{\hrule height.8pt depth0pt \kern2pt}%
  \def\@fs@post{}%
  \def\@fs@mid{\kern2pt\hrule\kern2pt}%
  \let\@fs@iftopcapt\iftrue}
\makeatother

\def\fullmodelname{Conditioned Location Diffusion}
\def\modelname{DiffTAD}

\iccvfinalcopy 

\begin{document}

\title{DiffTAD: Temporal Action Detection with Proposal Denoising Diffusion}

\author{Sauradip Nag$^{1,2}$ $\thanks{This work was done during internship with Jiankang Deng.}$
\and
Xiatian Zhu$^{1,3}$
\and
Jiankang Deng$^{4}$
\and
Yi-Zhe Song$^{1,2}$
\and 
Tao Xiang$^{1,2}$ 
\and \newline
{\small $^1$ CVSSP, University of Surrey, UK} ~ 
{\small $^2$ iFlyTek-Surrey Joint Research Center on Artificial Intelligence, UK} \\
{\small $^3$ Surrey Institute for People-Centred Artificial Intelligence, UK} ~
{\small $^4$ Imperial College London, UK} ~
}
\maketitle

\begin{abstract}
We propose a new formulation of temporal action detection (TAD) with denoising diffusion, \textbf{\em \modelname{}} in short.
Taking as input random temporal proposals,
it can yield action proposals accurately given an untrimmed long video.
This presents a generative modeling perspective,
against previous discriminative learning manners.
This capability is achieved by first diffusing the ground-truth proposals
to random ones (\ie, the forward/noising process) 
and then learning to reverse the noising process 
(\ie, the backward/denoising process).
Concretely, we establish the denoising process in the Transformer decoder (\eg, DETR) by introducing a temporal location query design
with faster convergence in training. 
We further propose a cross-step selective conditioning algorithm
for inference acceleration.
Extensive evaluations on ActivityNet and THUMOS show that our \modelname{} achieves top performance compared to previous art alternatives. The code will be made available at \href{https://github.com/sauradip/DiffusionTAD}{https://github.com/sauradip/DiffusionTAD}.
%

\end{abstract}

\section{Introduction}
Temporal action detection (TAD) aims to predict the temporal duration (\ie, start and end time) and the class label of each action instance in an untrimmed video \cite{idrees2017thumos,caba2015activitynet}. 
Existing methods 
rely on {\em proposal prediction} by regressing anchor proposals \cite{xu2017r,chao2018rethinking,gao2017turn,long2019gaussian}
or predicting the start/end times of proposals \cite{lin2019bmn,buch2017sst,lin2018bsn,xu2020g,nag2021few,xu2021boundary,xu2021low}.
These models are all discriminative learning based.

In the generative learning perspective, diffusion model \cite{song2021denoising} has been recently exploited in image based object detection \cite{chen2022diffusiondet}.
This represents a new direction for designing detection models in general.
Although conceptually similar to object detection,
the TAD problem presents more complexity due to 
the presence of temporal dynamics.
Besides, there are several limitations 
with the detection diffusion formulation in \cite{chen2022diffusiondet}.
First, a two-stage pipeline (\eg, RCNN \cite{chao2018rethinking}) is adopted, suffering localization-error propagation from proposal generation to proposal classification \cite{nag2022gsm}.
Second, as each proposal is processed individually, their relationship modeling is overlooked, potentially hurting the learning efficacy.
To avoid these issues, we adopt the one-stage detection pipelines \cite{tian2019fcos,wang2020solo}
that have already shown excellent performance with a relatively simpler design, in particular, DETR \cite{carion2020end}.



\begin{figure}
    \centering
    \includegraphics[scale=0.51]{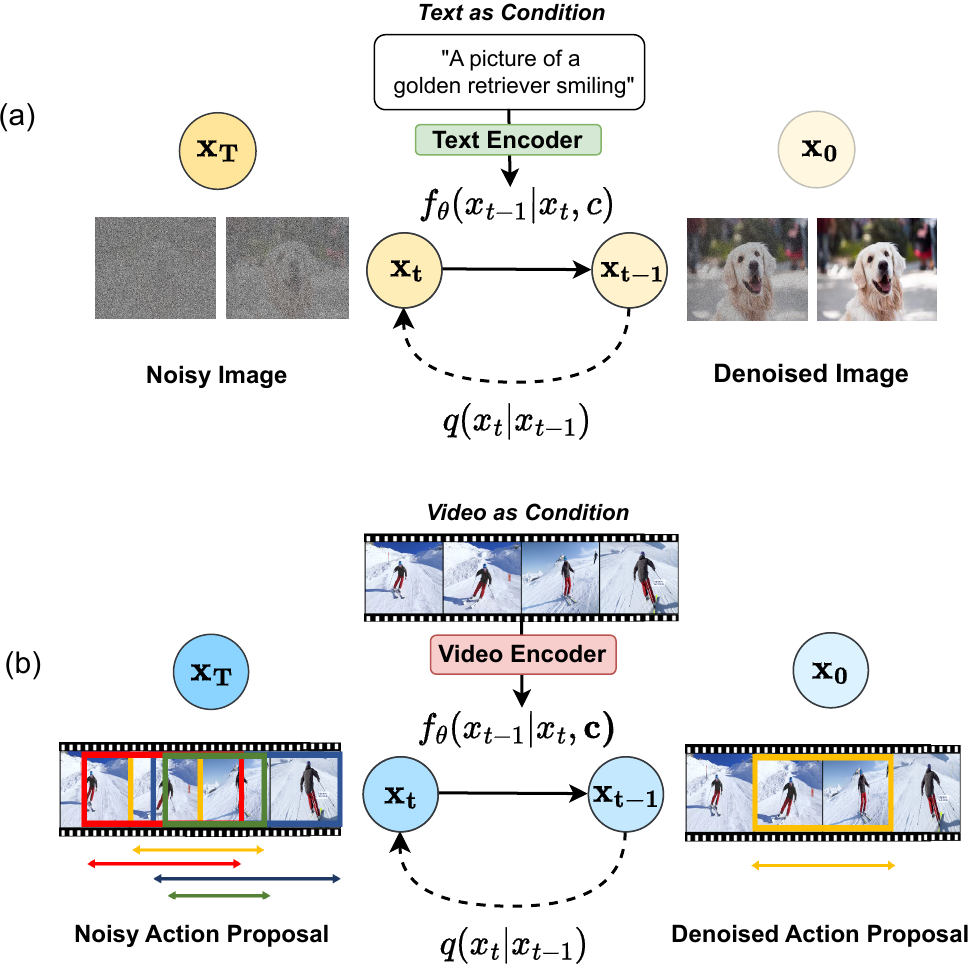} 
    \caption{\textbf{Diffusion for temporal action detection (TAD).} (a)
    A diffusion model for text-to-image generation where text embeddings are passed as the condition in the denoising (reverse) process. 
    We draw an analogy by exploiting 
    (b) a diffusion model for TAD: 
    {\em To generate action temporal boundaries from noisy proposals
    with the condition of video embedding.}
    }
    \vspace{-4mm}
    \label{fig:fig1}
\end{figure}

Nonetheless, it is non-trivial to integrate denoising diffusion with existing detection models, due to several reasons.
(1) 
Whilst efficient at handling high-dimension data simultaneously, diffusion models \cite{dhariwal2021diffusion,li2022diffusion} typically work with continuous input data.
But temporal locations in TAD are discrete.
(2) 
Denoising diffusion and action detection
both suffer low efficiency, and their combination would make it even worse.
Both of the problems have not been investigated systematically thus far.

To address the aforementioned challenges,
a novel {\bf\em \fullmodelname} method is proposed 
for efficiently tackling the TAD task in a diffusion formulation,
abbreviated as {\bf \modelname}.
It takes a set of random temporal proposals (\ie, the start and end time pairs) following Gaussian distribution,
and outputs the action proposals of a given untrimmed video.
At training time, Gaussian noises are first added to the ground truth proposals to make \emph{noisy} proposals. 
These discrete noisy proposals are then projected into a continuous vector space using sinusoidal projection \cite{liu2022dab} to form noisy queries in which the decoder (\eg, DETR) will conduct the denoising diffusion process. 
Our denoising space choice facilitates the adoption of existing diffusion models, as discussed above.
As a byproduct, the denoising queries strategy itself can accelerate the training convergence of DETR type models \cite{li2022dn}.
At inference time, conditioned on a test video, \modelname{} can generate action temporal boundaries by reversing the learned diffusion process
from Gaussian random proposals.
For improving inference efficiency,
we further introduce a cross-timestep selective conditioning mechanism 
with two key functions:
(1) minimizing the redundancy of intermediate predictions at each sampling step by filtering out the proposals far away from the distribution of corrupted proposals generated in training, and (2) conditioning the next sampling step by selected proposals 
to regulate the diffusion direction for more accurate inference.

Our \textbf{contributions} are summarized as follows. 
\noindent (1) For the first time we formulate the temporal action detection problem through denoising diffusion in the elegant transformer decoder framework. Additionally, integrating denoising diffusion with this decoder design
 solves the typical slow-convergence limitation.
\noindent (2) We further enhance the diffusion sampling efficiency and accuracy by introducing a novel selective conditioning mechanism during inference.
\noindent (3) Extensive experiments on ActivityNet and THUMOS benchmarks show that our \modelname{} achieves favorable performance against prior art alternatives.

\section{Related Works}

\noindent \textbf{Temporal action detection.} Inspired by object detection in static images \cite{ren2016faster},
R-C3D \cite{xu2017r} uses anchor proposals by following the design of proposal generation and classification.
With a similar model design, TURN \cite{gao2017turn} aggregates local features to represent snippet-level features for temporal boundary regression and classification. SSN \cite{zhao2017temporal} decomposes an action instance into three stages (starting, course, and ending)
and employs structured temporal pyramid pooling
to generate proposals.
BSN \cite{lin2018bsn} predicts the start, end, and actionness at each temporal location and generates proposals with high start and end probabilities. The actionness was further improved in BMN \cite{lin2019bmn} via
additionally generating a boundary-matching confidence map for improved proposal generation. 
GTAN \cite{long2019gaussian}
improves the proposal feature pooling procedure with a learnable Gaussian kernel for weighted averaging. G-TAD \cite{xu2020g}
learns semantic and temporal context via graph convolutional networks for more accurate proposal generation. BSN++ \cite{su2020bsn++} further extends BMN with a complementary boundary generator to capture rich context.
CSA \cite{sridhar2021class} enriches the proposal temporal context via attention transfer. VSGN \cite{zhao2021video} improves short-action localization using a cross-scale multi-level pyramidal architecture. \textcolor{black}{Recently, Actionformer \cite{zhang2022actionformer} and React \cite{shi2022react} proposed a purely DETR based design for temporal action localization at multiple scales.} 
Our DiffTAD is the very first TAD model which proposes action detection as a generative task.

\noindent \textbf{Diffusion models.} As a class of deep generative models, diffusion models~\cite{ho2020denoising, song2019generative, song2021scorebased} start from the sample in random distribution and recover the data sample via a gradual denoising process. 
Diffusion models have recently demonstrated remarkable results in fields including 
computer vision~\cite{avrahami2022blended, Ramesh2022HierarchicalTI, saharia2022photorealistic, pmlr-v162-nichol22a, gu2022vector, Fan2022FridoFP, Ruiz2022DreamBoothFT, singer2022make, harvey2022flexible, zhang2022motiondiffuse, ho2022video, yang2022diffusion},
nature language processing~\cite{austin2021structured, li2022diffusion, gong2022diffuseq},
audio processing~\cite{Popov2021GradTTSAD, yang2022diffsound, wu2021itotts, levkovitch2022zero, tae2021editts, huang2022prodiff, kim2022guided}, 
interdisciplinary applications\cite{jing2022torsional, hoogeboom2022equivariant, anand2022protein, xu2021geodiff, trippe2022diffusion, wu2022diffusion, arne2022structure}, \etc. More applications of diffusion models can be found in recent surveys~\cite{yang2022diffusion, cao2022survey}.

\begin{figure*}
    \centering
    \includegraphics[width=1.0\linewidth]
    {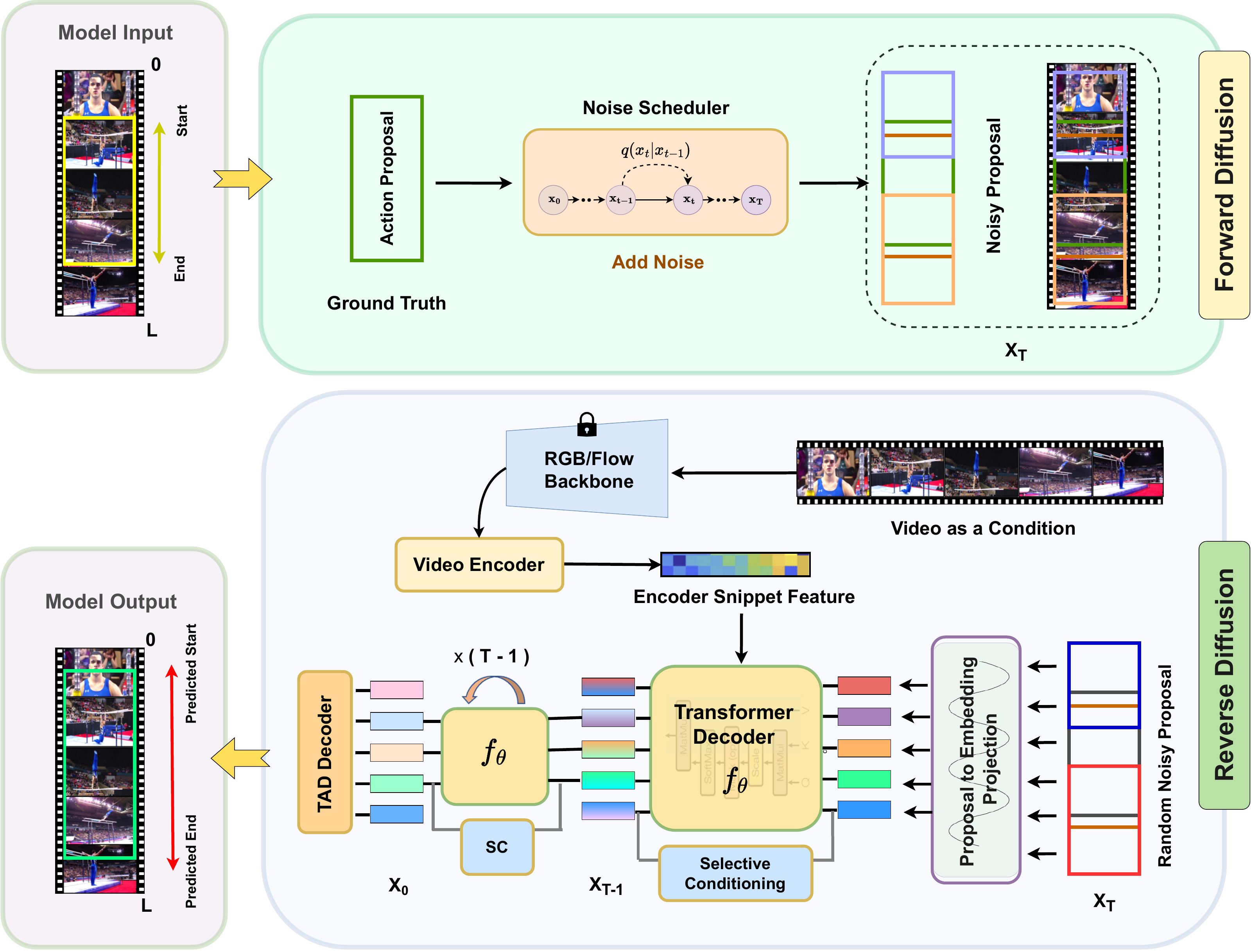}
    \caption{\textbf{Overview of our proposed DiffTAD.} In the forward diffusion process, Gaussian noises are added to the ground-truth boundaries iteratively to obtain noisy versions $X_{T}$. In the reverse denoising process, a video is passed as the condition along with random proposals sampled from Gaussian distribution. The discrete proposals are then projected to a continuous embedding space where proposal denoising takes place in an iterative fashion to obtain action proposals. In particular, a cross-timestep selective conditioning strategy is introduced for proposal refinement and filtering for more accurate and efficient inference.
     }
     \vspace{-4mm}
    \label{fig:my_label}
\end{figure*}

\noindent \textbf{Diffusion model for perception tasks.} While Diffusion models have achieved great success in image generation~\cite{ho2020denoising, song2021scorebased, dhariwal2021diffusion}, their potential for other perception tasks has yet to be fully explored. Some pioneer works tried to adopt the diffusion model for image segmentation tasks~\cite{wolleb2021diffusion, baranchuk2022labelefficient, graikos2022diffusion, kim2022diffusion, brempong2022denoising, amit2021segdiff, chen2022generalist}. For example, Chen~\etal~\cite{chen2022generalist} adopted Bit Diffusion model~\cite{chen2022analog} for panoptic segmentation~\cite{kirillov2019panoptic} of images and videos. However, despite significant interest in this idea,  
there is no previous solution that successfully adapts generative diffusion models for object detection, the progress of which remarkably lags behind that of segmentation. This might be because the segmentation task can be processed in an image-to-image style, which is more similar to image generation in formulation \cite{chen2023gss}. 
While object detection is a set prediction problem~\cite{carion2020end} with a need for assigning object candidates~\cite{ren2015faster,lin2017feature,carion2020end} to ground truth objects. However, Chen \etal~\cite{chen2022diffusiondet} managed to apply a diffusion model to object detection for the first time. 
Similarly, we make the first attempt at formulating TAD
in the diffusion framework by integrating the denoising process with 
the single-stage DETR architecture.

\section{Methodology}
\subsection{Preliminaries}

\noindent \textbf{Temporal action detection.} Our \modelname{} model takes as input an untrimmed video $V$ with a variable number of frames. Video frames are first pre-processed by a feature encoder (\eg, a Kinetics pre-trained I3D network \cite{carreira2017quo}) into a sequence of localized snippets following the standard practice \cite{lin2019bmn}. To train the model, we collect a set of labeled video training set $\mathcal{D}^{train} = \{V_i, \Psi_i\}$. Each video $V_i$ is labeled with temporal annotation $\Psi_i = \{(\psi_j, \xi_j, y_j)\}_{j=1}^{M_i}$ where $\psi_{j}$/$\xi_{j}$ denote the start/end time, $y_j$ is the action category, and $M_i$ is the number of action instances. 

\noindent \textbf{Diffusion models}~\cite{sohl2015deep, ho2020denoising, song2019generative, song2021denoising} are a class of likelihood-based models inspired by nonequilibrium thermodynamics~\cite{song2019generative, song2020improved}. These models define a Markovian chain of diffusion forward process by gradually adding noises to the sample data. The forward noising process is defined as
\begin{equation}
\label{eq:noise_process}
    q(\bm{z}_t | \bm{z}_0) = \mathcal{N}(\bm{z}_t | \sqrt{\bar{\alpha}_t} \bm{z}_0, (1 - \bar{\alpha}_t) \bm{I}),
\end{equation}
which transforms a sample $\bm{z}_0$ to a latent noisy sample $\bm{z}_t$ ($t\in\{0, 1, ...,T\}$) by adding noises to $\bm{z}_0$.
$\bar{\alpha}_t \coloneqq \prod_{s=0}^{t} \alpha_s = \prod_{s=0}^{t} (1 - \beta_s)$ and $\beta_s$ represents the noise variance schedule~\cite{ho2020denoising}.
During training, a neural network $f_\theta(\bm{z}_t, t)$ is trained to predict $\bm{z}_0$ from $\bm{z}_t$ by minimizing the training objective with $\ell_2$ loss~\cite{ho2020denoising}:
\begin{equation}
    \mathcal{L}_\text{train} =  \frac{1}{2}|| f_\theta(\bm{z}_t, t) - \bm{z}_0 ||^2.
\end{equation}
At inference, a sample $\bm{z}_0$ is reconstructed from noise $\bm{z}_T$ with the model $f_\theta$ and an updating rule~\cite{ho2020denoising, song2021denoising} in an iterative way, \ie,  $\bm{z}_T \rightarrow \bm{z}_{T-\Delta} \rightarrow ... \rightarrow \bm{z}_0$. More details of diffusion models can be found in Supplementary.


\subsection{\modelname}
\label{sec:difftad}
\paragraph{Diffusion-based TAD formulation.}
In this work, we formulate the temporal action detection task in 
a conditional denoising diffusion framework. In our setting, data samples are a set of action temporal boundaries $\bm{z}_0 = \bm{b}$, where $\bm{b}\in \mathbb{R}^{N \times 2}$ denotes $N$ temporal proposals. 
A neural network $f_\theta(\bm{z}_t, t, \bm{x})$ is trained to predict $\bm{z}_0$ from noisy proposals $\bm{z}_t$, conditioned on the corresponding video $\bm{V}$. The corresponding category label $\bm{\hat{y}}$ is produced accordingly. 

Since the diffusion model generates a data sample iteratively, it needs to run the model $f_\theta$ multiple times in inference. However, it would be computationally intractable to directly apply $f_\theta$ on the raw video at every iterative step.
For efficiency, we propose to separate the whole model into two parts, \textit{video encoder} and \textit{detection decoder}, where the former runs only once to extract a feature representation of the input video $\bm{V_{i}}$, and the latter takes this feature as a condition
to progressively refine the 
noisy proposals $\bm{z}_t$.

\noindent \textbf{Video encoder.} 
The video encoder takes as input the 
pre-extracted video features and extracts high-level features for the following detection decoder. In general, any video encoder can be used. 
We use a video encoder same as \cite{zhang2022actionformer}.
%
More specifically, the raw video $V$ is first 
encoded by a convolution encoder to obtain multiscale feature $H_{i} \in \mathbb{R}^{T \times D}$ for RGB and optical flow separately. This is followed by a multi-scale temporal transformer \cite{vaswani2017attention} $\mathcal{T}$ that performs global attention across the time dimension to obtain the global feature as:
\begin{align}
    F^{i}_{g} = \mathcal{T}(H_{i}) , i \in [1,\cdots,L] 
\end{align}
where query/key/value of the transformer is set to $H_{i}$ and $L$ is the number of scales. We estimate the shared global representations $F^{i}_{g} \in \mathbb{R}^{T \times D}$ across all the scales and concatenate them as $F_{g} = \{ F^{0}_{g}, F^{1}_{g}, ..., F^{L}_{g}\}$. 

\begin{figure}[t]
    \centering
    \includegraphics[width=1.0\linewidth]{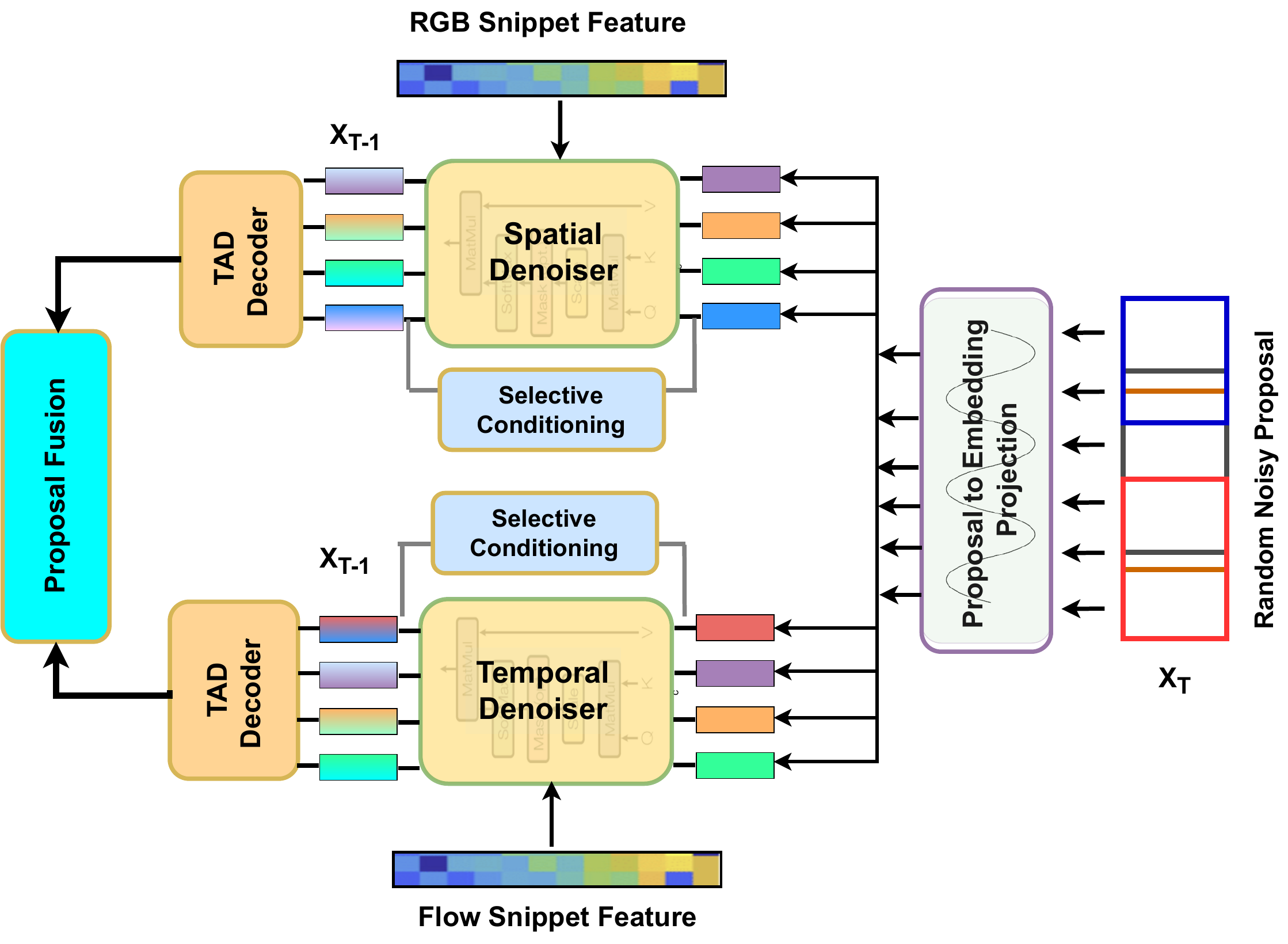} 
    \caption{\textbf{Spacetime decoupled denoising.} We perform spacetime decoupled denoising via feature late-fusion. The RGB and optical flow features are extracted separately and passed as a condition to the denoiser, separately.
    }
    \label{fig:decoup}
\end{figure}

Previous TAD works \cite{lin2019bmn,xu2020g} typically 
use fused RGB and flow features (\ie, early fusion)
for modeling.
%
Considering the feature specificity (RGB for appearance, and optical flow for motion), 
we instead exploit a late fusion strategy (Fig.~\ref{fig:decoup}).
Specifically, we extract the video features $F^{rgb}_{g}$ and $F^{flow}_{g}$ for RGB and optical flow separately.
The proposal denoising is then conducted in each space individually.
The forward/noising process is random (\ie, feature agnostic) and thus
shared by both features.
We will compare the two fusion strategies empirically
(see Table~\ref{tab:fusion}).


\noindent \textbf{Detection decoder.} Similar to DETR \cite{lin2021detr}, we use a transformer decoder \cite{vaswani2017attention} (denoted by $f_{\theta}$) for detection.
Functionally it serves as a denoiser. 
In traditional DETR, the queries are learnable continuous embeddings with random initialization. 
%
%
In \modelname{}, however, we exploit the queries as the {\em denoising targets}.
%

Specifically, we first project discrete proposals $\psi \in \mathbb{R}^{N \times 2}$ to continuous query embeddings \cite{wang2021anchor}:
\begin{align}
    Q = g(\psi) \in \mathbb{R}^{N \times D}
\end{align}
where $g$ is MLP-based learnable projection.
Taking $Q$ as input, 
the decoder then predicts $N$ outputs:
\begin{align}
    F_{d} = f_{\theta}(Q;F_{g}) \in \mathbb{R}^{N \times D}
\end{align}
where $F_{g}$ is the global encoder feature and 
the $F_{d}$ is the final embedding.
$F_{d}$ is finally decoded using three parallel heads namely (1) \textit{action classification head}, (2) \textit{action localization head}, and (3) \textit{completeness head}, respectively. The first estimates the probability of a particular action within the action proposal. The second estimates the start, end and IOU overlap between the proposals. The third estimates the quality of predicted action proposals.




\paragraph{Cross-timestep selective conditioning.}
In \modelname{}, the denoising decoder $f_{\theta}$ takes $N$ action queries and then denoises each of them iteratively.
Processing a large number of queries is thus inefficient.
An intuitive way for better efficiency is to use static thresholds to suppress unimportant candidates \cite{chen2022diffusiondet}, which however is ad-hoc and ineffective.

\begin{figure}[t]
    \centering
    \includegraphics[scale=0.25]{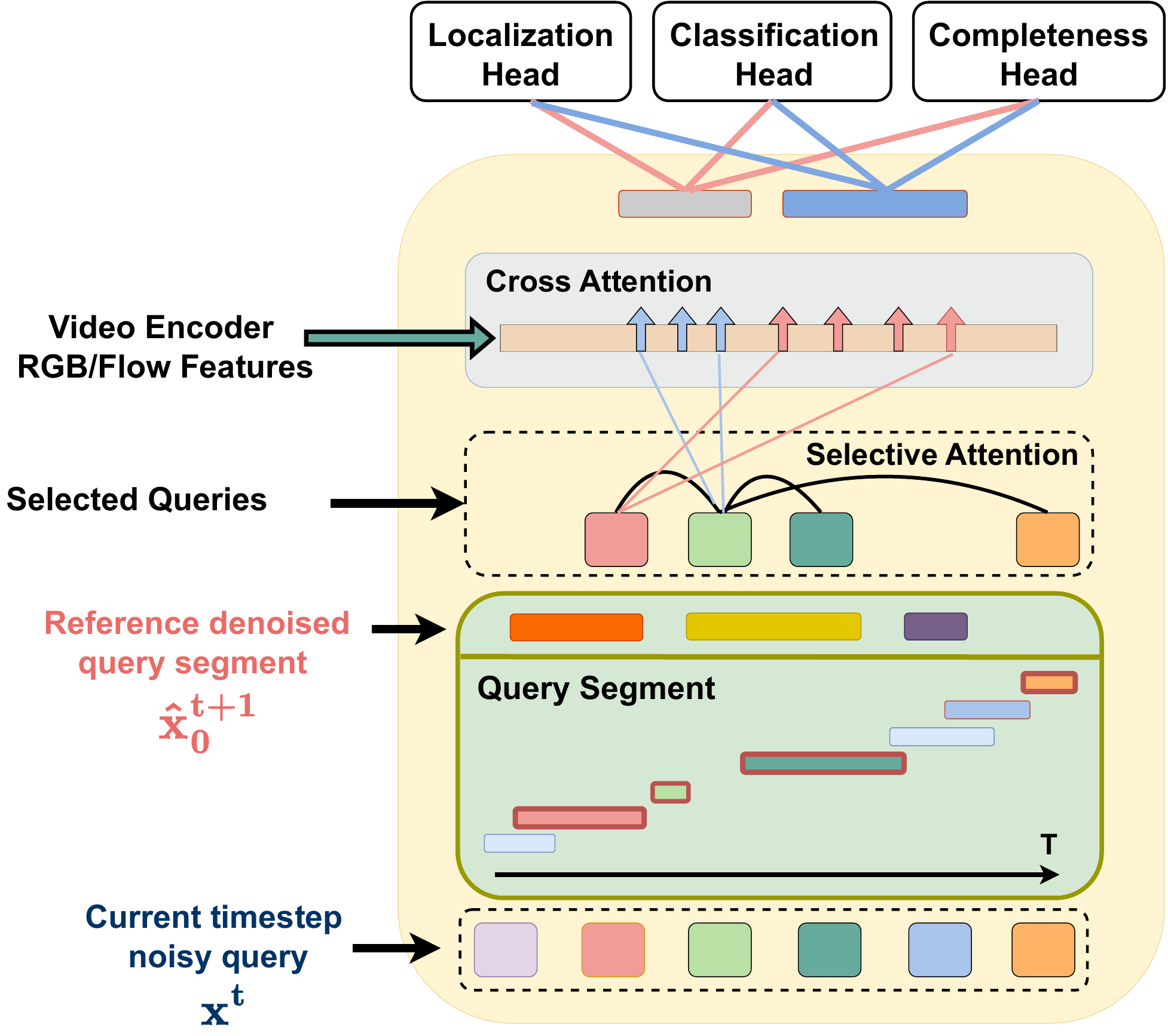} 
    \caption{\textbf{Cross-timestep selective conditioning.} At the current time step, noisy query only attends to other queries selectively based on the overlap and similarity with the previously denoised reference proposals/segments. For clarity, LayerNorm, FFN, and residual connection are omitted.}
    \label{fig:sel}
\end{figure}

Here, we propose a more principled {\em cross-timestep selective conditioning} mechanism (Fig. \ref{fig:sel}). 
More specifically, 
we calculate a similarity matrix $A \in \mathbb{R}^{N \times N}$ between the $N$ queries $x_{t}$ of current timestep and $N$ queries $x_{t+1}$ of conditioned/previous timestep $x_{t+1}$.
Each element of $A$ represents the similarity of the same queries across two successive timesteps.
We select a set of queries according to: 
\begin{align}
    \hat{P} _{sim} = \{(i,j)|A[i,j] - \gamma > 0 \}
\end{align}
where $\gamma \in [-1,1]$ is a preset similarity threshold. 
We consider higher IoU with the desired (approximated by the estimate of the last step) proposals,
more effective for the queries to be denoised.
%
Thus, we construct an IoU based matrix $B \in \mathbb{R}^{N \times N}$ between two successive timesteps:
\begin{align}
    \hat{P}_{iou} = \{(i,j)|B[i,j] - \gamma > 0 \}
\end{align}
where $i$/$j$ indexes the queries.
This allows for the most useful queries to be selected (see Fig. 1 in Supplementary).

We obtain the 
final query set as $Q_{c} = (\hat{P}_{iou} / \hat{P}_{sim}) \bigcup Q$ with $/$ denotes the set divide operation. 
For a selected query $q_{i}$, its key and value features can be obtained by fusion as $K_{i}/V_{i} = cat(\{k_{j}/v_{j}|(i,j) \in Q_{c}\})$. Each query feature is then updated as 
\begin{align}
    \hat{q}_{i} = softmax(q_{i}K^{\top}_{i}).V^{\top}_{i}
\end{align}
These selected queries $\{\hat{q}_{i}\}$
will be passed through the cross-attention with video embeddings for denoising.
This selective conditioning is applied on
a random proportion (\eg, 70\%) of proposals per mini-batch during training, and on all the proposals during inference.


Our selective conditioning strategy shares the spirit of self-conditioning \cite{chen2022analog} in the sense that the output of the last step is used for accelerating the subsequent denoising.
However, our design and problem setup both are different.
For example, self-conditioning can be simply realized
by per-sample feature concatenation, whilst our objective
is to select a subset of queries
based on the pairwise similarity and IoU measurement in an attention manner.
We validate the conditioning design in the experiments 
(Table \ref{tab:selcond}).

\subsection{Training}
During training, we first construct the diffusion process that corrupts the ground-truth proposals to noisy proposals.
We then train the model to reverse this noising process. Please refer to Algorithm~\ref{alg:train} for more details.

\begin{algorithm}[h]
\small
\caption{\small DiffTAD Training 
}
\label{alg:train}
\algcomment{\fontsize{7.2pt}{0em}\selectfont \texttt{alpha\_cumprod(t)}: cumulative product of $\alpha_i$, \ie, $\prod_{i=1}^t \alpha_i$
}
\definecolor{codeblue}{rgb}{0.25,0.5,0.5}
\definecolor{codegreen}{rgb}{0,0.6,0}
\definecolor{codebluenew}{RGB}{52, 70, 235}
\definecolor{codekw}{RGB}{207,33,46}
\lstset{
  backgroundcolor=\color{white},
  basicstyle=\fontsize{7.5pt}{7.5pt}\ttfamily\selectfont,
  columns=fullflexible,
  breaklines=true,
  captionpos=b,
  commentstyle=\fontsize{7.5pt}{7.5pt}\color{codebluenew},
  keywordstyle=\fontsize{7.5pt}{7.5pt}\color{codekw},
  escapechar={|}, 
}
\begin{lstlisting}[language=python]
def train(video_feat, gt_proposals):

  # Encode image features
  feats = video_encoder(video_feat)

  # Signal scaling
  pb = (pb * 2 - 1) * scale  

  # Corrupt gt_proposals
  t = randint(0, T)|~~~~~~~~~~~|# time step
  eps = normal(mean=0, std=1)  # noise: [B, N, 2]
  
  pb_crpt = sqrt(|~~~~|alpha_cumprod(t)) * pb + 
              |~|sqrt(1 - alpha_cumprod(t)) * eps

  # Project to continuous embedding
  pb_crpt = project(pb_crpt) # query : [B, N, D]
              
  # Calculate Self-condition estimate
  pb_pred = zeros_like(pb_crpt)
  if self_cond and uniform(0,1) > 0.7:
    pb_pred = decoder(pb_crpt, pb_pred, feats, t)
    pb_pred = stop_gradient(pb_pred)
  
  # Predict
  pb_pred = decoder(pb_crpt, pb_pred, feats, t)

  # Set prediction loss
  loss = set_prediction_loss(pb_pred, gt_proposals)
  
  return loss
\end{lstlisting}
\end{algorithm}

\paragraph{Proposal corruption.}
We add Gaussian noises to the ground truth action proposals. The noise scale is controlled by $\alpha_{t}$~(in Eq.~\eqref{eq:noise_process}), which adopts the monotonically decreasing cosine schedule in different timestep $t$, following~\cite{nichol2021improved}. Notably, the ground truth proposal coordinates need to be scaled as well since the signal-to-noise ratio has a significant effect on the performance of diffusion model~\cite{chen2022generalist}. We observe that TAD favors a relatively lower signal scaling value than object detection \cite{chen2022diffusiondet} (see Table \ref{tab:snr}).
More discussions are given in Supplementary.

\paragraph{Training losses.}
The detection detector takes as input $N_{train}$ corrupted proposals and predicts $N_{train}$ predictions each including the category classification, proposal coordinates, and IOU regression. We apply the set prediction loss~\cite{carion2020end, sun2021sparse, zhu2021deformable} on the set of $N_{train}$ predictions. We assign multiple predictions to each ground truth by selecting the top $k$ predictions with the least cost by an optimal transport assignment method~\cite{ge2021ota, ge2021yolox, wu2022defense, du20211st}. 

\subsection{Inference}\label{sec:inference}

In inference, starting from noisy proposals sampled in Gaussian distribution, the model progressively refines the predictions as illustrated in Algorithm~\ref{alg:sample}.

\begin{algorithm}[t]
    \small
\caption{\small DiffTAD Sampling 
}
\label{alg:sample}
\algcomment{\fontsize{7.2pt}{0em}\selectfont \texttt{linespace}: generate evenly spaced values
}
\definecolor{codeblue}{rgb}{0.25,0.5,0.5}
\definecolor{codegreen}{rgb}{0,0.6,0}
\definecolor{codebluenew}{RGB}{52, 70, 235}
\definecolor{codekw}{rgb}{0.85, 0.18, 0.50}
\lstset{
  backgroundcolor=\color{white},
  basicstyle=\fontsize{7.5pt}{7.5pt}\ttfamily\selectfont,
  columns=fullflexible,
  breaklines=true,
  captionpos=b,
  commentstyle=\fontsize{7.5pt}{7.5pt}\color{codebluenew},
  keywordstyle=\fontsize{7.5pt}{7.5pt}\color{codekw},
  escapechar={|}, 
}
\begin{lstlisting}[language=python]
def infer(video_feat, steps, T):
  
  # Encode video features
  feats = video_encoder(video_feat)

  # noisy proposals: [B, N, 2]
  pb_t = normal(mean=0, std=1)

  # noisy embeddings: [B, N, D]
  pb_t = project(pb_t) 
  
  pb_pred = zeros_like(pb_t)

  # uniform sample step size
  times = reversed(linespace(-1, T, steps))
  
  # [(T-1, T-2), (T-2, T-3), ..., (1, 0), (0, -1)]
  time_pairs = list(zip(times[:-1], times[1:])

  for t_now, t_next in zip(time_pairs):
    # Predict pb_0 from pb_t
    if not self_cond:
        pb_pred = zeros_like(pb_t)
    pb_pred = decoder(pb_t, pb_pred, feats, t_now)
    
    # Estimate pb_t at t_next
    pb_t = ddim_step(pb_t, pb_pred, t_now, t_next)
 
  return pb_pred
\end{lstlisting}

\end{algorithm}



\paragraph{Sampling step.}
At each sampling step, the random or estimated proposals from the last sampling step are first projected into the continuous query embeddings and sent into the detection decoder to predict the category, proposal IOU and proposal coordinates. After obtaining the proposals of the current step, DDIM~\cite{song2021denoising} is adopted to estimate the proposals for the next step. 

\paragraph{Proposal prediction.}  
DiffTAD has a simple proposal generation pipeline without post-processing (\eg, non-maximum suppression).
To make a reliable confidence estimation for each proposal, we fuse the action classification
score $p_{bc}$ and completeness score $p_{c}$ for each proposal with
a simple average to obtain the final proposal score $p_{sc}$. 
%
%

\subsection{Remarks}

\begin{figure}[t]
    \centering
    \includegraphics[scale=0.26]{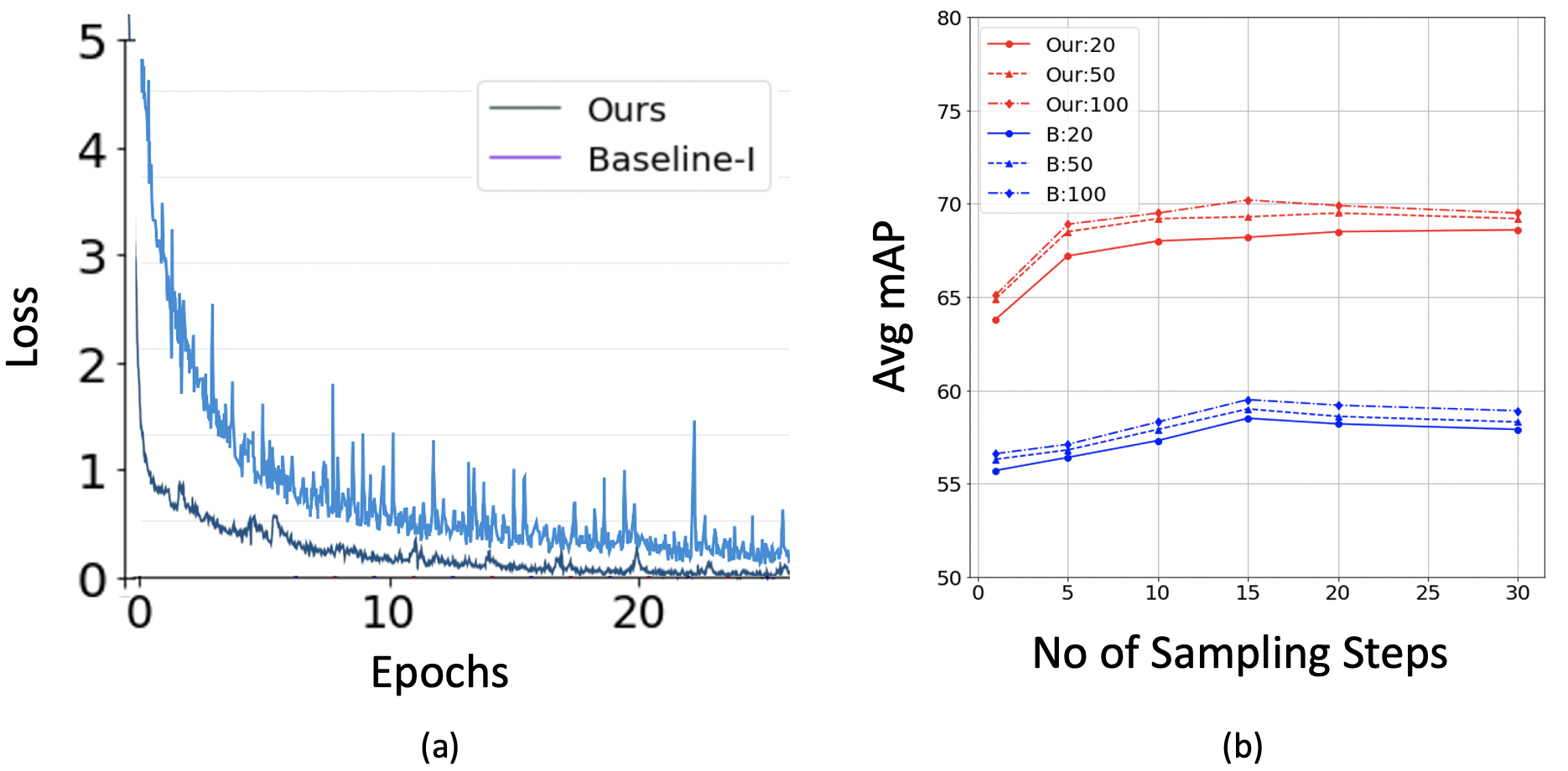} 
    \caption{\textbf{Properties of \modelname.}
    (a) Our proposed model converges more stably during training. 
    (b) Its performance increases with the number of queries (20/50/100) and also with the sampling steps, with a clear margin over the Baseline (B-20/50/100).
    Dataset: THUMOS.}
    \label{fig:properties}
\end{figure}

\paragraph{One model multiple trade-offs.}
Once trained,
\modelname{} works under multiple settings with a varying number of proposals and sampling steps during inference (Fig.~\ref{fig:properties}(b)). 
In general, better accuracy can be obtained using more proposals and more steps.
Thus, a single \modelname{} can realize a number of trade-off needs between speed and accuracy.


\paragraph{Faster convergence.}
DETR variants suffer generally slow convergence \cite{liu2022dabdetr} due to two reasons.
First, inconsistent updates of the anchors, the objective for the object queries to learn, would make the optimization of target boundaries difficult.
Second, the ground truth assignment using a dynamic process (\eg, Hungarian matching) is unstable due to both the nature of discrete bipartite matching and the stochastic training process. For instance, a small perturbation with the cost matrix might cause enormous matching and inconsistent optimization.
Our \modelname{} takes a denoising strategy 
that makes learning easier.
More specifically, each query is designed as a proposal proxy, a noised query, that can be regarded as a good anchor due to being close to a ground truth. 
%
%
With the ground truth proposal as a definite optimization objective,
the ambiguity brought by Hungarian matching can be well suppressed. 
We validate that our query denoising based \modelname{} converges more stably than DiffusionDet \cite{chen2022diffusiondet} based Baseline-I (Fig.~\ref{fig:properties}(a)), whilst achieving superior performance (Table \ref{tab:sota}).



\paragraph{Better sampling.}
We evaluate \modelname{} under a varying number of (10/50/100) random proposals by increasing the sampling steps from 1 to 30.
As shown in Table~\ref{tab:sota}, under all three settings, \modelname{} yields steady performance gains from more steps consumed. In particular, in the case of fewer random proposals, often \modelname{} can achieve larger gains than DiffusionDet \cite{chen2022diffusiondet}.
For example, in the case of 50 proposals,
the mAP of \modelname{} boosts the avg mAP from 64.9\%~(1 step) to 68.5\%~(5 steps), \ie, an absolute gain of 3.6\% avg mAP. 
Unlike object detection, we find TAD benefits little from increasing the number of proposals.
One intuitive reason is that positive samples are less in TAD than in object detection. 
Compared to the previous two-stage refinement of discriminative learning based TAD models \cite{tan2021relaxed}, our gain is also more decent ({{0.8\% {\em vs.} 4.2\% (10 steps)}}).
This is because it lacks a principled iterative inference ability as in diffusion models.

\begin{table*}[t]
\caption{Performance comparison with the state-of-the-art methods.
Metrics: mAP at different IoU thresholds, and average mAP in {[}0.3 : 0.1 : 0.7{]} on THUMOS14 and {[}0.5 : 0.05 : 0.95{]} on ActivityNet-v1.3.}
\label{tab:sota}
\centering
\begin{tabular}{clccccccccccc}
\hline
\multicolumn{1}{c|}{}                                        & \multicolumn{1}{l|}{}                                  & \multicolumn{1}{c|}{}                                   & \multicolumn{6}{c|}{\textbf{THUMOS}}                                                                                                                                                                                  & \multicolumn{4}{c}{\textbf{ActivityNet}}                                                                                               \\ \cline{4-13} 
\multicolumn{1}{c|}{\multirow{-2}{*}{\textbf{Models}}}       & \multicolumn{1}{l|}{\multirow{-2}{*}{\textbf{Design}}} & \multicolumn{1}{c|}{\multirow{-2}{*}{\textbf{Feature}}} & \textbf{0.3}               & \textbf{0.4}               & \textbf{0.5}               & \textbf{0.6}               & \multicolumn{1}{c|}{\textbf{0.7}}               & \multicolumn{1}{c|}{\textbf{Avg}}               & \textbf{0.5}               & \textbf{0.75}              & \multicolumn{1}{c|}{\textbf{0.95}}              & \textbf{Avg}               \\ \hline
\multicolumn{13}{c}{\bf \em Discriminative learning based models}                                                                                                                                                                                                                                                                                                                                                                                                                                                                                                              \\ \hline
\multicolumn{1}{c|}{TAL-Net \cite{chao2018rethinking}}                                     & \multicolumn{1}{l|}{2-stage}                           & \multicolumn{1}{c|}{I3D}                                &  53.2                       & 48.5                       & 42.8                        & 33.8                       & \multicolumn{1}{c|}{20.8}                       & \multicolumn{1}{c|}{-}                       &  38.2                       & 18.3 & \multicolumn{1}{c|}{ 1.3}                        & 20.2                       \\
\multicolumn{1}{c|}{BMN \cite{lin2019bmn}}                                     & \multicolumn{1}{l|}{2-stage}                           & \multicolumn{1}{c|}{TSN}                                & 56.0                       & 47.4                       & 38.8                       & 29.7                       & \multicolumn{1}{c|}{20.5}                       & \multicolumn{1}{c|}{38.5}                       & 50.1                       & 34.8                       & \multicolumn{1}{c|}{8.3}                        & 33.9                       \\
\multicolumn{1}{c|}{GTAD \cite{xu2020g}}                                    & \multicolumn{1}{l|}{2-stage}                           & \multicolumn{1}{c|}{TSN}                                & 54.5                       & 47.6                       & 40.3                       & 30.8                       & \multicolumn{1}{c|}{23.4}                       & \multicolumn{1}{c|}{39.3}                       & 50.4                       & 34.6                       & \multicolumn{1}{c|}{9.0}                        & 34.1                       \\
\multicolumn{1}{c|}{RTD-Net \cite{tan2021relaxed}}                                 & \multicolumn{1}{l|}{2-stage}                           & \multicolumn{1}{c|}{I3D}                                & 68.3                       & 62.3                       & 51.9                       & 38.8                       & \multicolumn{1}{c|}{23.7}                       & \multicolumn{1}{c|}{-}                          & 47.2                       & 30.7                       & \multicolumn{1}{c|}{8.6}                        & 30.8                       \\
\multicolumn{1}{c|}{TCANet \cite{qing2021temporal}}                                  & \multicolumn{1}{l|}{2-stage}                           & \multicolumn{1}{c|}{I3D}                                & 60.6                       & 53.2                       & 44.6                       & 36.8                       & \multicolumn{1}{c|}{26.7}                       & \multicolumn{1}{c|}{44.3}                       & 52.3                       & 36.7                       & \multicolumn{1}{c|}{6.9}                        & 35.5                       \\
\multicolumn{1}{c|}{MUSES \cite{liu2021multi}}                                   & \multicolumn{1}{l|}{2-stage}                           & \multicolumn{1}{c|}{I3D}                                & 68.9                       & 64.0                       & 56.9                       & 46.3                       & \multicolumn{1}{c|}{31.0}                       & \multicolumn{1}{c|}{—}                          & 50.0                       & 35.0                       & \multicolumn{1}{c|}{6.6}                        & 34.0                       \\
\multicolumn{1}{c|}{ContextLoc \cite{zhu2021enriching}}                              & \multicolumn{1}{l|}{2-stage}                           & \multicolumn{1}{c|}{I3D}                                & 68.3                       & 63.8                       & 54.3                       & 41.8                       & \multicolumn{1}{c|}{26.2}                       & \multicolumn{1}{c|}{50.9}                       & 56.0                       & 35.2                       & \multicolumn{1}{c|}{3.6}                        & 34.2                       \\
\multicolumn{1}{c|}{RCL \cite{wang2022rcl}}                                    & \multicolumn{1}{l|}{2-stage}                           & \multicolumn{1}{c|}{I3D}                                & 70.1                        & 62.3                        & 52.9                        & 42.7                        & \multicolumn{1}{c|}{30.7}                       & \multicolumn{1}{c|}{57.1}                       & 51.7                         & 35.2                       & \multicolumn{1}{c|}{8.0}                        & 34.4                       \\
\multicolumn{1}{c|}{React \cite{shi2022react}}                                   & \multicolumn{1}{l|}{1-stage}                           & \multicolumn{1}{c|}{I3D}                                & 69.2                       & 65.0                       & 57.1                       & 47.8                       & \multicolumn{1}{c|}{35.6}                       & \multicolumn{1}{c|}{55.0}                       & 49.6                       & 33.0                       & \multicolumn{1}{c|}{8.6}                        & 32.6                       \\
\multicolumn{1}{c|}{TAGS \cite{nag2022gsm}}                                   & \multicolumn{1}{l|}{1-stage}                           & \multicolumn{1}{c|}{I3D}                                & 68.6                        & 63.8                        & 57.0                        & 46.3                        & \multicolumn{1}{c|}{31.8 }                       & \multicolumn{1}{c|}{52.8 }                       & \bf{56.3}                        & 36.8                        & \multicolumn{1}{c|}{\bf{9.6} }                        & \bf{36.5}                       \\ 
\multicolumn{1}{c|}{ActionFormer \cite{zhang2022actionformer}}                            & \multicolumn{1}{l|}{1-stage}                           & \multicolumn{1}{c|}{I3D}                                & 82.1                       & 77.8                       & 71.0                       & 59.4                       & \multicolumn{1}{c|}{43.9}                       & \multicolumn{1}{c|}{66.8}                       & 53.5                       & 36.2                       & \multicolumn{1}{c|}{8.2}                        & 35.6                       \\ \hline
\multicolumn{13}{c}{\bf\em Generative learning based models}                                                                                                                                                                                                                                                                                                                                                                                                                                                                                                                  \\ \hline
\multicolumn{1}{c|}{Baseline(1-step)}                                & \multicolumn{1}{l|}{2-stage}                           & \multicolumn{1}{c|}{I3D}                                & 65.2                          & 61.3                          & 55.4                          & 44.6                         & \multicolumn{1}{c|}{35.5}                         & \multicolumn{1}{c|}{52.4}                         & 48.5                         & 31.4                         & \multicolumn{1}{c|}{8.6}                         & 31.5                         \\

\multicolumn{1}{c|}{Baseline(5-step)}                                & \multicolumn{1}{l|}{2-stage}                           & \multicolumn{1}{c|}{I3D}                                & 69.1                         & 65.7                         & 60.2                         & 47.1                         & \multicolumn{1}{c|}{36.4}                         & \multicolumn{1}{c|}{55.7}                         & 50.2                         & 32.3                         & \multicolumn{1}{c|}{8.9}                         & 32.2                         \\
\multicolumn{1}{c|}{Baseline(10-step)}                                & \multicolumn{1}{l|}{2-stage}                           & \multicolumn{1}{c|}{I3D}                                & 70.0                         & 66.5                         & 60.6                         & 47.5                         & \multicolumn{1}{c|}{36.9}                         & \multicolumn{1}{c|}{56.3}                         & 51.0                         & 32.9                         & \multicolumn{1}{c|}{9.0}                         & 32.4                         \\
\hline
\multicolumn{1}{c|}{\cellcolor[HTML]{CBCEFB} {\bf DiffTAD}(1-step)} & \multicolumn{1}{l|}{\cellcolor[HTML]{CBCEFB} 1-stage}                           & \multicolumn{1}{c|}{\cellcolor[HTML]{CBCEFB}I3D}        & \cellcolor[HTML]{CBCEFB}68.7 & \cellcolor[HTML]{CBCEFB}66.8 & \cellcolor[HTML]{CBCEFB}64.7 & \cellcolor[HTML]{CBCEFB}\bf{61.2} & \multicolumn{1}{c|}{\cellcolor[HTML]{CBCEFB}\bf{57.3}} & \multicolumn{1}{c|}{\cellcolor[HTML]{CBCEFB}63.8} & \cellcolor[HTML]{CBCEFB}52.4 & \cellcolor[HTML]{CBCEFB}35.6 & \multicolumn{1}{c|}{\cellcolor[HTML]{CBCEFB}8.8} & \cellcolor[HTML]{CBCEFB}34.8 \\ 
\multicolumn{1}{c|}{\cellcolor[HTML]{CBCEFB}{\bf DiffTAD}(5-step)} & \multicolumn{1}{l|}{\cellcolor[HTML]{CBCEFB} 1-stage}                           & \multicolumn{1}{c|}{\cellcolor[HTML]{CBCEFB}I3D}        & \cellcolor[HTML]{CBCEFB}73.4 & \cellcolor[HTML]{CBCEFB}71.5 & \cellcolor[HTML]{CBCEFB}69.9 & \cellcolor[HTML]{CBCEFB}\bf{62.8} & \multicolumn{1}{c|}{\cellcolor[HTML]{CBCEFB}\bf{58.4}} & \multicolumn{1}{c|}{\cellcolor[HTML]{CBCEFB}\bf{67.2}} & \cellcolor[HTML]{CBCEFB}55.2 & \cellcolor[HTML]{CBCEFB}\bf{36.8} & \multicolumn{1}{c|}{\cellcolor[HTML]{CBCEFB}8.9} & \cellcolor[HTML]{CBCEFB}36.0 \\
\multicolumn{1}{c|}{\cellcolor[HTML]{CBCEFB}{\bf DiffTAD}(10-step)} & \multicolumn{1}{l|}{\cellcolor[HTML]{CBCEFB} 1-stage}                           & \multicolumn{1}{c|}{\cellcolor[HTML]{CBCEFB}I3D}        & \cellcolor[HTML]{CBCEFB}74.9 & \cellcolor[HTML]{CBCEFB}72.8 & \cellcolor[HTML]{CBCEFB}\bf{71.2} & \cellcolor[HTML]{CBCEFB}\bf{62.9} & \multicolumn{1}{c|}{\cellcolor[HTML]{CBCEFB}\bf{58.5}} & \multicolumn{1}{c|}{\cellcolor[HTML]{CBCEFB}\bf{68.0}} & \cellcolor[HTML]{CBCEFB}56.1 & \cellcolor[HTML]{CBCEFB}\bf{36.9} & \multicolumn{1}{c|}{\cellcolor[HTML]{CBCEFB}9.0} & \cellcolor[HTML]{CBCEFB}36.1 \\\hline
\end{tabular}
\end{table*}

\section{Experiments}
\noindent{\bf Datasets.}
We conduct extensive experiments on two popular TAD benchmarks.
(1) \textit{ActivityNet-v1.3}~\cite{caba2015activitynet} has 19,994 videos from 200 action classes. We follow the standard setting 
to split all videos into training, validation and testing subsets 
in ratio of 2:1:1.
(2) \textit{THUMOS14}~\cite{idrees2017thumos} has 200 validation videos and 213 testing videos from 20 categories with labeled temporal boundary and action class.

\subsection{Implementation details}

\noindent \textbf{Training schedule.}
%
For video feature extraction, we use
Kinetics pre-trained I3D model \cite{carreira2017quo,zhang2022actionformer} with a downsampling ratio of 4 and R(2+1)D model \cite{alwassel2020tsp,zhang2022actionformer} with a downsampling ratio of 2. Our model is trained for 50 epochs using Adam optimizer with a learning rate of {$10^{-4}/10^{-5}$ for AcitivityNet/THUMOS respectively. The batch size is set to 40 for ActivityNet and 32 for THUMOS}.
For selective conditioning, we apply the rate of 70\%
during training. 
All models are trained with 4 NVIDIA-V100 GPUs. 

\noindent \textbf{Testing schedule.}
At the inference stage, the detection decoder iteratively refines the predictions from Gaussian random proposals. By default, we set the total sampling time steps as 10. 

\subsection{Main results}
\paragraph{Competitors.} 
We compare our \modelname{} with the state-of-the-art non-generative approaches including BMN \cite{lin2019bmn}, GTAD \cite{xu2020g}, React \cite{shi2022react} and ActionFormer \cite{zhang2022actionformer}. 
Further, we adapt the object detection counterpart DiffusionDet \cite{chen2022diffusiondet} 
to a two-stage generative TAD method, termed as \texttt{Baseline}. 

\paragraph{Results on THUMOS.}
We make several observations from Table \ref{tab:sota}:
(1) Our {\modelname} achieves the best result,
surpassing strong discriminative learning based competitors like TCANet \cite{wang2021temporal} and ActionFormer \cite{zhang2022actionformer} by a clear margin. 
This suggests the overall performance advantage of our model design and generative formulation idea.
(2) Importantly, \modelname{} excels on the alternative generative learning design (\ie, \texttt{Baseline}),
validating the superiority of our diffusion-based detection formulation.
For both generative models, more sampling steps lead to higher performance.
This concurs with the general property of diffusion models.
(3) In particular, our model achieves significantly stronger results in stricter IOU metrics (\eg, IOU@0.5/0.6/0.7), as compared to all the other alternatives. This demonstrates the potential of generative learning in tackling the action boundary often with high ambiguity, and the significance of proper diffusion design.

\paragraph{Results on ActivityNet.}
Similar observations can be drawn in general on ActivityNet from Table \ref{tab:sota}.
We further highlight several differences:
(1) Indeed, overall our \modelname{} is not the best performer, with a slight edge underneath TAGS \cite{nag2022gsm}.
However, we note that all other DETR style methods (\eg, RTD-Net) are significantly inferior. 
This means that our method has already successfully filled up the most performance disadvantage of the DETR family.
We attribute this result to our design choice of denoising in the query space and cross-timestep selective conditioning.
%
That being said, our formulation in exploiting the DETR architecture for TAD
is superior than all prior attempts.
(2) 
Compared to the generative competitor (\ie, \texttt{Baseline}),
our model is not only more stable to converge (Fig. \ref{fig:properties}(a)),
but also yields a margin of 3.7\%
(smaller than that on THUMOS as this is a more challenging test for the DETR family in general).
%

\subsection{Ablation study}
We conduct ablation experiments on THUMOS to study \modelname{} in detail. 
In all experiments, 
we use 30 proposals for training and inference,
unless specified otherwise.

\begin{figure}
    \centering
    \includegraphics[scale=0.238]{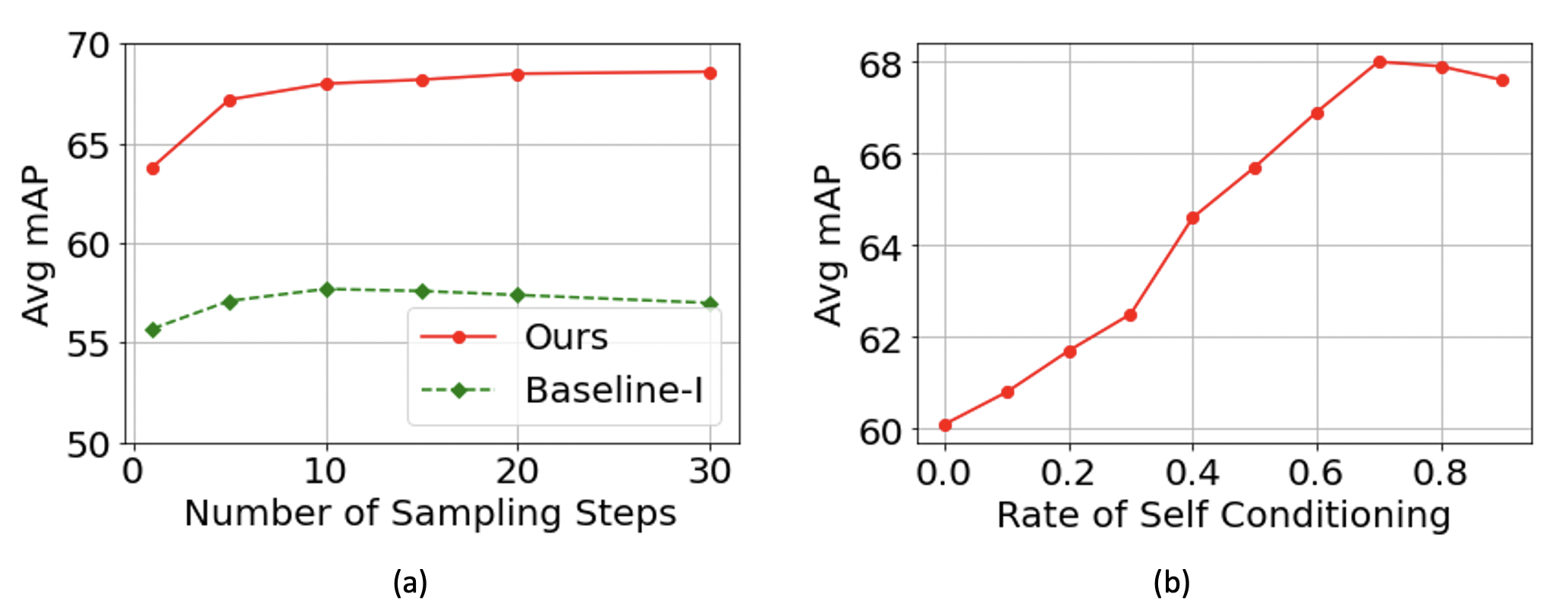} 
    \caption{\textbf{Impact of sampling and selective-conditioning.} (a) The effect of varying sampling steps with Baseline and \modelname{}.
    (b) The effect of selective conditioning rate during training \modelname{}.
    Dataset: THUMOS.
    }
    \label{fig:selfc}
\end{figure}

\paragraph{Cross-timestep selective conditioning.} 
We examine the effect of the proposed selective conditioning
for proposal refinement (Section \ref{sec:difftad}).
To that end, we vary the rate/portion of proposals per batch at which selective conditioning is applied during training.
The case of 0\% training rate means {\em no conditioning}.
Note, during inference selective conditioning is always applied to all the proposals (\ie, 100\% test rate).
As demonstrated in Fig.~\ref{fig:selfc}(b), we observe a clear correlation between conditioning rate and sampling quality (\ie, mAP), validating the importance of our selective conditioning in refining proposals for enhanced denoising.

Additionally, we compare with two different refinement designs:
(1) {\em Feature concatenation} as in self-conditioning \cite{chen2022analog},
and
(2) {\em Proposal renewal} as in DiffusionDet \cite{chen2022diffusiondet}.
We observe from Table~\ref{tab:selcond} that our selective conditioning is superior to both alternatives,
verifying our design idea.





\begin{table}[t]
\centering
\caption{\textbf{Proposal refinement design.} 
Dataset: THUMOS.
}
\label{tab:selcond}
\begin{tabular}{c|cccc}
\hline
                                             & \multicolumn{4}{c}{\textbf{mAP}}                                                                      \\ \cline{2-5} 
\multirow{-2}{*}{\textbf{Design}} & \textbf{0.3} & \textbf{0.5} & \multicolumn{1}{c|}{\textbf{0.7}}                        & \textbf{Avg} \\ \hline
Feature Concatenation \cite{chen2022analog}                                & 71.1           & 66.3           & \multicolumn{1}{c|}{56.5}                                  & 65.9           \\
Proposal Renewal \cite{chen2022diffusiondet}                                & 71.2           & 65.4           & \multicolumn{1}{c|}{54.2}                                  & 65.1
\\ 
\rowcolor[HTML]{EFEFEF} 
\textbf{Selective Condition} (Ours)                          & \textbf{74.9}  & \textbf{71.2}  & \multicolumn{1}{c|}{\cellcolor[HTML]{EFEFEF}\textbf{58.5}} & \textbf{68.0}  \\ \hline
\end{tabular}
\end{table}

\paragraph{Sampling decomposition.}
We decompose the sampling strategy with \modelname{}
by testing four variants:
(1) {\em No denoising}: No DDIM \cite{song2021denoising} is applied, which means 
the output prediction of the current step is used directly as the input of the next step (\ie, a naive baseline).
(2) {\em Vanilla denoising}: DDIM is applied in the original form.
(3) {\em Selective conditioning}:
Only selective conditioning is used but no vanilla denoising.
(4) {\em Both vanilla denoising and selective conditioning}:
Our full model.
We observe from Table~\ref{tab:samp} that
(1) Without proper denoising, applying more steps 
will cause more degradation, as expected,
because there is no optimized Markovian chain. 
(2) Applying vanilla denoising can address the above problem
and improve the performance from multiple sampling accordingly.
However, the performance increase saturates quickly.
(3) Interestingly, our proposed selective conditioning 
even turns out to be more effective than vanilla denoising.
(4) When both vanilla denoising and selective conditioning are applied (\ie, our full model), the best performances can be achieved, with a clear gain over the either and even improved saturation phenomenon. This suggests that the two ingredients are largely complementary, which is not surprising given their distinctive design nature.  

%
%
%

\begin{table}[t]
\centering
\caption{\textbf{Sampling decomposition} in terms of iterative denoising ({\bf ID}) and selective conditioning ({\bf SC}). Dataset: THUMOS.
}
\label{tab:samp}
\begin{tabular}{c|c|c|c|c}
\hline
\textbf{ID} & \textbf{SC} & \textbf{step 1} & \textbf{step 5} & \textbf{step 10} \\ \hline
\xmark            & \xmark                    & 62.7              & 62.9              & 61.3              \\
\cmark           & \xmark                    & 62.7              & 65.0              & 64.9              \\
\xmark            & \cmark                   & 62.7              & 65.7              & 65.6              \\ \hline
\rowcolor[HTML]{EFEFEF} 
\cmark           & \cmark                   & 62.7              &\bf 67.2              &\bf 68.0              \\ \hline
\end{tabular}
\end{table}

\paragraph{Signal scaling.}
The signal scaling factor (Eq.~\eqref{eq:noise_process}) controls the signal-to-noise ratio~(SNR) of the denoising process. 
We study the influence of this facet in \modelname{}.
As shown in Table~\ref{tab:snr}, the scaling factor of 0.5 achieves the optimal performance.
This means this setting is task-specific at large (\eg, 
the best choice is 2.0 for object detection 
\cite{chen2022diffusiondet}).
%

\begin{table}[t]
\centering
\caption{\textbf{Signal-noise ratio} under a variety of different 
scaling factors. Dataset: THUMOS.
}
\label{tab:snr}
\begin{tabular}{c|cccc}
\hline
                                 & \multicolumn{4}{c}{\textbf{mAP}}                                                                      \\ \cline{2-5} 
\multirow{-2}{*}{\textbf{Scale}} & \textbf{0.3} & \textbf{0.5} & \multicolumn{1}{c|}{\textbf{0.7}}                        & \textbf{Avg} \\ \hline
0.1                              & 70.5           & 67.5           & \multicolumn{1}{c|}{57.2}                                  & 66.6           \\
\rowcolor[HTML]{EFEFEF} 
\textbf{0.5}                     & \textbf{74.9}  & \textbf{71.2}  & \multicolumn{1}{c|}{\cellcolor[HTML]{EFEFEF}\textbf{58.5}} & \textbf{68.0}  \\
1.0                              & 73.2           & 70.6           & \multicolumn{1}{c|}{58.0}                                  & 67.3           \\
2.0                              & 69.9           & 67.3           & \multicolumn{1}{c|}{56.8}                                  & 64.2           \\ \hline
\end{tabular}
\end{table}

\paragraph{Accuracy vs. speed.}
We evaluate the trade-off between the accuracy and memory cost with  \modelname{}. 
In this test, we experiment with three choices $\{10, 50,100\}$ in the number of proposals.
For each choice, the same is applied to both model training and evaluation consistently.
We observe in Table~\ref{tab:speed} that:
(1) Increasing the number of proposals/queries from 30 to 
50 brings about 0.5\% in mAP with 
extra 0.46 
GFLOPs, thanks to the proposed selective conditioning.
(2) However, further increasing the proposals is detrimental to the model performance. This is because the number of action instances per video is much fewer than that of objects per image on average.
(3) With 50 proposals, increasing the sampling steps from 1 to 5 provides an mAP gain of $3.6\%$ with 
an additional cost of 1.9 GFLOPs.
A similar observation is with the case of 100 proposals.


\begin{table}[]
\centering
\small
\caption{\textbf{Accuracy vs. speed} under a variety of different proposal numbers and sampling steps.
Dataset: THUMOS.}
\label{tab:speed}
\begin{tabular}{c|c|cccc|c}
\hline
                                   &                                 & \multicolumn{4}{c|}{\textbf{mAP}}                                                                     &                                  \\ \cline{3-6}
\multirow{-2}{*}{\textbf{Proposals}} & \multirow{-2}{*}{\textbf{Step}} & \textbf{0.3} & \textbf{0.5} & \multicolumn{1}{c|}{\textbf{0.7}}                        & \textbf{Avg} & \multirow{-2}{*}{\textbf{GFLOPs}} \\ \hline
\cellcolor[HTML]{EFEFEF}30                                 & \cellcolor[HTML]{EFEFEF}5                               & \cellcolor[HTML]{EFEFEF}74.9           & \cellcolor[HTML]{EFEFEF}71.2           & \multicolumn{1}{c|}{\cellcolor[HTML]{EFEFEF}58.5}                                  & \cellcolor[HTML]{EFEFEF}68.0           & \cellcolor[HTML]{EFEFEF}0.81                              \\

50                                & 1                               & 69.2           & 65.2           & \multicolumn{1}{c|}{53.0}                                  & 64.9           & 1.27                               \\

\textbf{50}                       & \textbf{5}                      & \textbf{77.2}  & \textbf{73.5}  & \multicolumn{1}{c|}{\textbf{59.1}} & \textbf{68.5}  & \textbf{3.17}                      \\
100                                & 1                               & 70.4           & 66.1           & \multicolumn{1}{c|}{53.8}                                  & 65.1           & 2.18                               \\
100                                & 5                               & 77.1           & 73.8           & \multicolumn{1}{c|}{58.7}                                  & 68.4           & 6.32                               \\
\hline
\end{tabular}
\end{table}

\begin{figure}[t]
    \centering
    \includegraphics[width=1.0\linewidth]{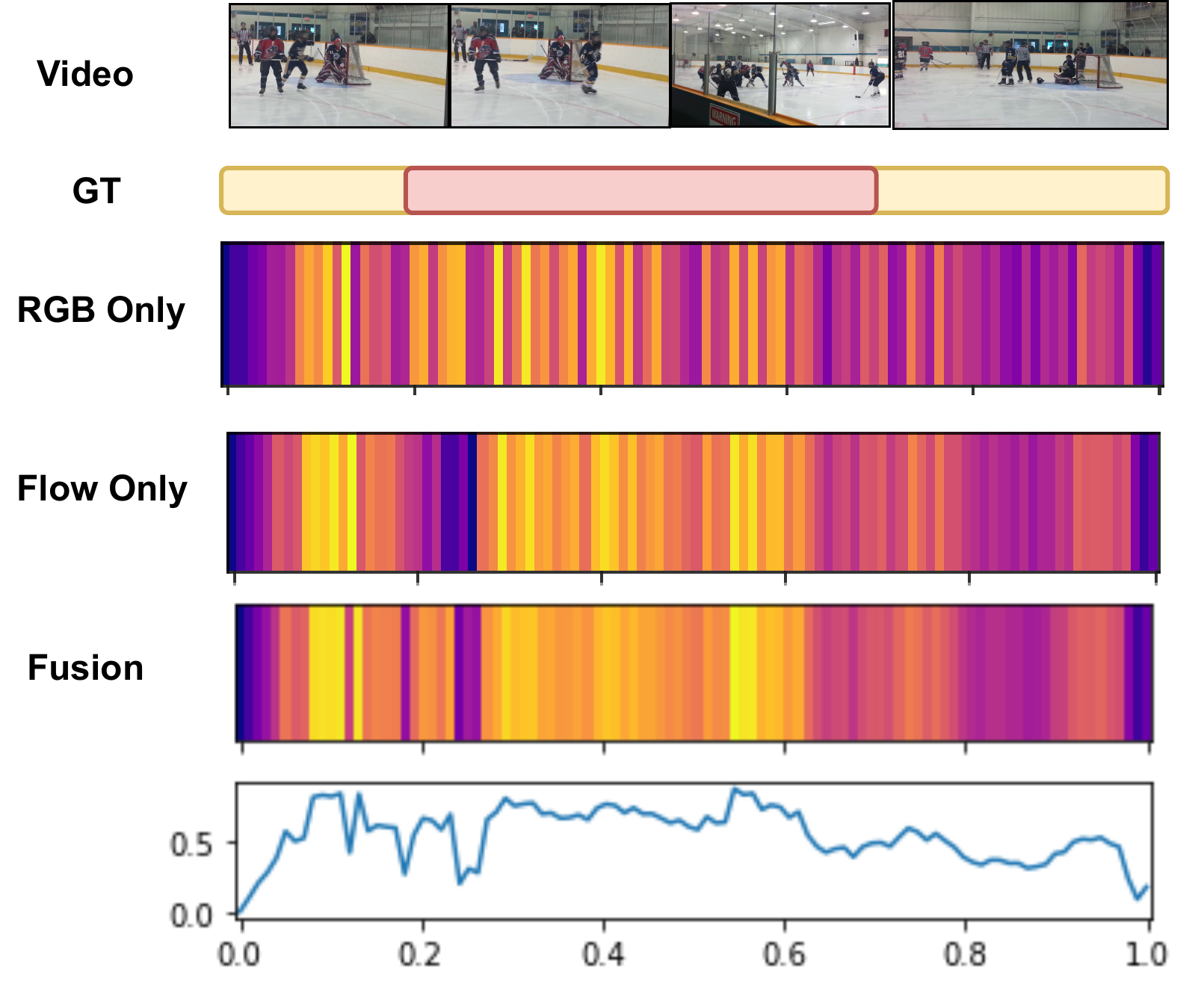} 
     \caption{\textbf{Late fusion example} on a test video from THUMOS. 
     It is observed that RGB and optical flow features are complementarily useful in capturing the ground truth (GT) action instance.}
    \label{fig:fusion}
\end{figure}

\paragraph{Decoupled diffusion strategy.} 
We evaluate the impact of feature decoupling (video encoder in Sec. \ref{sec:difftad}). 
Typically, existing TAD methods \cite{xu2020g,lin2019bmn}
use fused RGB and optical flow video features (\ie, early fusion).
However, we take a late fusion strategy where the RGB and flow features are processed separately before their proposals are fused 
(Fig. \ref{fig:decoup}).
We contrast the two fusion strategies.
It is evident in Table~\ref{tab:fusion}
that with the typical early fusion (\ie, passing early fused features as a condition to our detection decoder),
a drop of $2\%$ in average mAP is resulted.
This indicates that due to modal specificity 
there is a need for specific conditional inference,
validating our design consideration.
%
For visual understanding, an example is given in Fig.~\ref{fig:fusion} to show how the two features can contribute to TAD in a cooperative manner.

\begin{table}[t]
\centering
\caption{\textbf{Decoupling the denoising.}
RGB and optical flow video features are decoupled for individual denoising (\ie, late fusion),
in contrast to typical early fusion strategy where
the two features are first fused in prior to being processed.
Dataset: THUMOS.
}
\label{tab:fusion}
\begin{tabular}{c|ccc|c}
\hline
\textbf{Decoupling} & \textbf{0.5} & \textbf{0.75} & \textbf{0.95} & \textbf{Avg} \\ \hline
\xmark{}                 & 71.8           & 67.4            & 57.1            & 66.0           \\ \hline
\rowcolor[HTML]{C0C0C0} 
\textbf{\cmark{}}            & \textbf{74.9}  & \textbf{71.2}   & \textbf{58.5}   & \textbf{68.0}  \\ \hline
\end{tabular}
\end{table}

\paragraph{NMS-free design.} 
As shown in Table~\ref{tab:snms}, when comparing DiffTAD with and without non-maximum suppression (NMS), we observe similar results. NMS is not necessary in DiffTAD because the predictions are relatively
sparse and minor overlapped with our cross-step conditional strategy (Sec.~\ref{sec:difftad}). In contrast, existing non-generative TAD works like BMN \cite{lin2019bmn} and GTAD \cite{xu2020g} generate highly overlapped proposals with similar confidence. Thus, NMS becomes necessary
to suppress redundant proposals. 
\begin{table}[t]
\centering
\caption{Ablation study on non-maximum suppression (NMS) on THUMOS14, measured by AR@AN.}
\label{tab:snms}
\begin{tabular}{c|c|c|c}
\hline
\textbf{Model} & \textbf{AR@50} & \textbf{AR@100} & \textbf{AR@500} \\ \hline 
BMN \cite{lin2019bmn}       &  29.04           & 37.72           & 56.07   \\ 
BMN \cite{lin2019bmn}  + NMS       &  32.73           & 40.68           & 56.42   \\ \hline
DiffTAD & 63.6  & 69.6   & \textbf{73.1} \\ 
DiffTAD + NMS & \textbf{64.3}  & \textbf{69.9}   & 72.8            \\ \hline
\end{tabular}
\end{table}

\paragraph{Ablation of denoising strategy.} Due to the inherent query based design with the detection decoder, (1) we can corrupt discrete action proposals and project them as queries,
and (2) we can also corrupt the action label and project it as label queries.
Both can be taken as the input to the decoder. 
For corrupting the label queries, we use random shuffle as the noise in the forward diffusion step.
To validate this design choice experimentally, we test three variants: (1) only labels are corrupted,
(2) only proposals are corrupted, and
(3) both proposals and labels are corrupted. 
For the non-corrupted quantity,
we add noise to the randomly initialized embedding. 
For the last variant, we stack all the corrupted proposals and labels and pass them into the decoder. 
It can be observed in Table~\ref{tab:denoise} that corrupting  labels alone observes the most drop in performance, and corrupting both labels and proposals is inferior to only corrupting the proposals.

\begin{table}[t]
\centering
\caption{Denoising strategy with DiffTAD on THUMOS14.}
\label{tab:denoise}
\begin{tabular}{cc|cccc}
\hline
\multicolumn{2}{c|}{Denoising Strategy}                            & \multicolumn{4}{c}{mAP}                                           \\ \hline
\multicolumn{1}{c|}{Proposal}               & Label & 0.3 & 0.5 & \multicolumn{1}{c|}{0.7}                        & Avg \\ \hline
\multicolumn{1}{c|}{\xmark}                          & \cmark             & 70.1  & 65.9  & \multicolumn{1}{c|}{52.7}                         & 62.4  \\
\rowcolor[HTML]{EFEFEF} 
\multicolumn{1}{c|}{\cellcolor[HTML]{EFEFEF}\cmark} & \xmark              & 74.9  & 71.2  & \multicolumn{1}{c|}{\cellcolor[HTML]{EFEFEF}58.5} & 68.0  \\
\multicolumn{1}{c|}{\cmark}                         & \cmark             & 73.8  & 70.4  & \multicolumn{1}{c|}{58.0}                         & 67.3  \\ \hline
\end{tabular}
\end{table}

\begin{figure*}[t]
    \centering
    \includegraphics[scale=0.185]{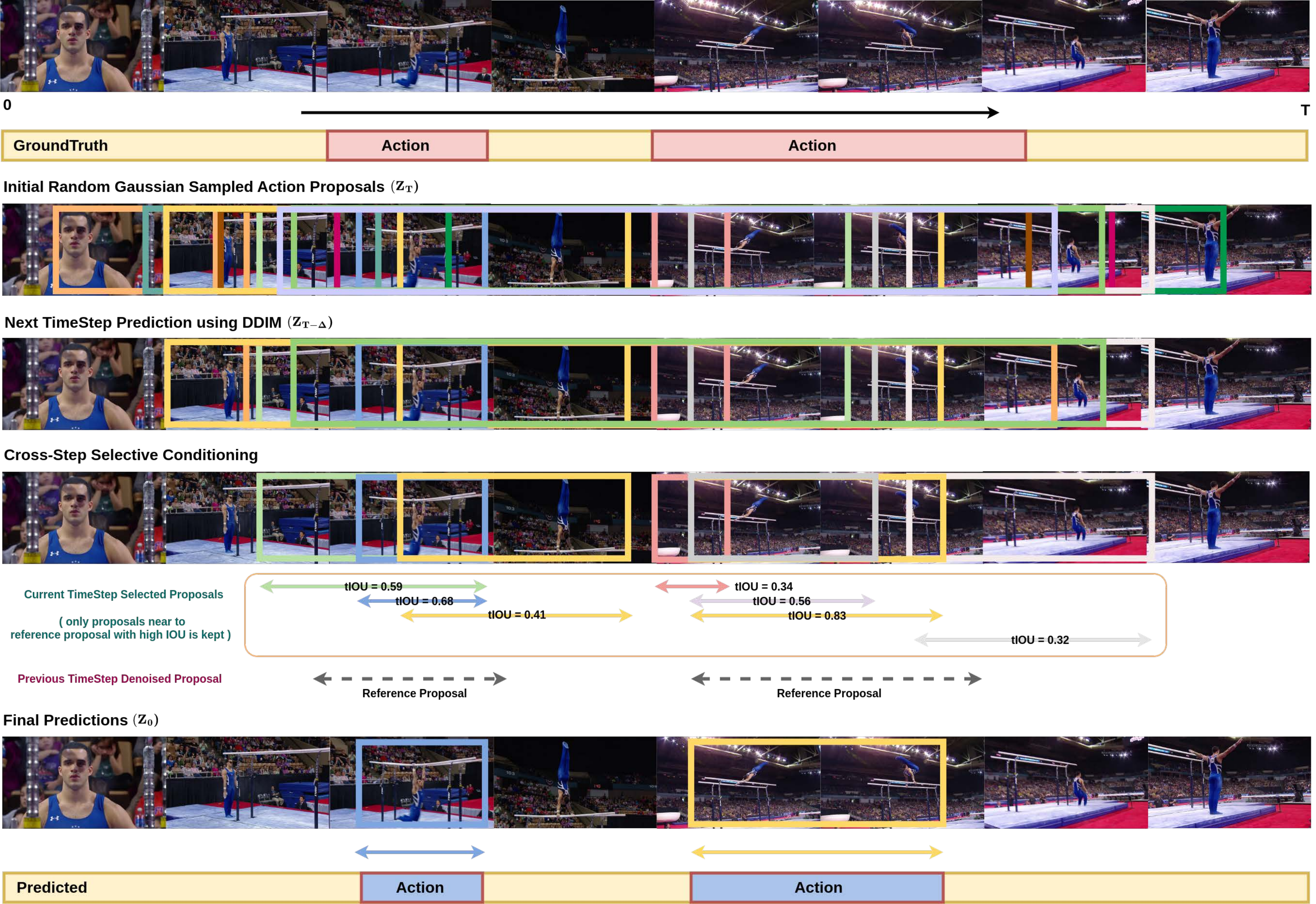}
    \caption{\textbf{Visualization of DiffTAD} proposal denoising step during inference}
    \label{fig:visual}
\end{figure*}

\section{Visualization of proposal denoising}

We visualize our sampling step of DiffTAD in Fig~\ref{fig:visual}. The model is initialized with 30 proposals for better visualization. This experiment is performed on a random testing video from ActivityNet dataset. 

\textit{(a)} Initial action proposals are randomly sampled from the Gaussian distribution ($Z_{T}$) and then projected as queries into the detection decoder $f_{\theta}$. 

\textit{(b)} The detection decoder predicts the action proposals (start/end point) along with the action class. The noise is calculated and then the action proposals are denoised using DDIM based denoising diffusion strategy. 

\textit{(c)} Our proposed cross-step selection strategy estimates the best candidate proposals based on the denoised reference proposal from the last step. The proposals with low temporal overlap with the reference proposals are dropped from the denoising step thus accelerating the inference. 

\textit{(d)} After multiple steps of refinement, final denoised action proposal predictions are obtained.

\section{Conclusion}

In this work, we propose a novel temporal action detection (TAD) paradigm, \modelname{}, by considering it as a denoising diffusion process from noisy proposals to action proposals. 
This proposed generative model is conceptually distinctive from all previous TAD methods based on discriminative learning.
Our model is designed by properly tailoring the diffusion and denoising process in a single-stage DETR framework,
with appealing properties such as more stable convergence, flexible proposal sizes, and superior proposal refinement.
%
Experiments on standard benchmarks show that \modelname{} achieves favorable performance compared to both generative and non-generative alternatives. 



{\small
\bibliographystyle{ieee_fullname}
\bibliography{main_arxiv}
}

\end{document}